\documentclass[10pt,twocolumn,letterpaper]{article}

\usepackage{cvpr}              

\usepackage{graphicx}
\usepackage{amsmath}
\usepackage{amssymb}
\usepackage{multirow}
\usepackage{booktabs}
\usepackage{colortbl}  
\usepackage{xcolor}
\usepackage{array}  
\usepackage{booktabs}
\usepackage[accsupp]{axessibility}

\definecolor{cvprblue}{rgb}{0.21,0.49,0.74}
\usepackage[pagebackref,breaklinks,colorlinks,allcolors=cvprblue]{hyperref}

\def\confName{CVPR}
\def\confYear{2026}

\usepackage{tcolorbox}
\usepackage{etoolbox}
\usepackage{graphicx}
\usepackage{amsmath}
\usepackage[normalem]{ulem}
\usepackage{amssymb}
\usepackage{multirow}
\usepackage{booktabs}
\usepackage{colortbl}  
\usepackage{xcolor}
\usepackage{array}  
\usepackage{booktabs}
\usepackage[accsupp]{axessibility}

\usepackage[table,dvipsnames]{xcolor}
\definecolor{gray}{rgb}{0.9, 0.9, 0.9} 
\usepackage{amsmath}
\DeclareMathOperator*{\argmax}{arg\,max}

\definecolor{cvprblue}{rgb}{0.21,0.49,0.74}
\newcommand{\norm}[1]{\left\lVert#1\right\rVert}

\title{ Language-driven Fine-grained Retrieval}

\author{
Shijie Wang, Xin Yu, Yadan Luo, Zijian Wang, Pengfei Zhang, Zi Huang \\
   \textsuperscript{} The University of Queensland, Austrilia \\}

\begin{document}
\maketitle
\begin{abstract}
Existing fine-grained image retrieval (FGIR) methods learn discriminative embeddings by adopting semantically sparse one-hot labels derived from category names as supervision. 
While effective on seen classes, such supervision overlooks the rich semantics encoded in category names, hindering the modeling of comparability among cross-category details and, in turn, limiting generalization to unseen categories.
To tackle this, we introduce LaFG, a \uline{\textbf{La}}nguage-driven framework for \uline{\textbf{F}}ine-\uline{\textbf{G}}rained Retrieval that converts class names into attribute-level supervision using large language models (LLMs) and vision–language models (VLMs). 
Treating each name as a semantic anchor, LaFG prompts an LLM to generate detailed, attribute-oriented descriptions. 
To mitigate attribute omission in these descriptions, it leverages a frozen VLM to project them into a vision-aligned space, clustering them into a dataset-wide attribute vocabulary while harvesting complementary attributes from related categories.
Leveraging this vocabulary, a global prompt template selects category-relevant attributes, which are aggregated into category-specific linguistic prototypes. 
These prototypes supervise the retrieval model to steer it toward pinpointing visual details consistent with linguistic descriptions, thus modeling comparability among object details. 
Extensive evaluations show that LaFG achieves impressive performance on both fine- and coarse-grained benchmarks and generalizes well to unseen categories. 

\end{abstract}

\section{Introduction}

\begin{figure}
     \centering
     \begin{subfigure}[b]{1\linewidth}
         \centering
         \includegraphics[width=\linewidth]{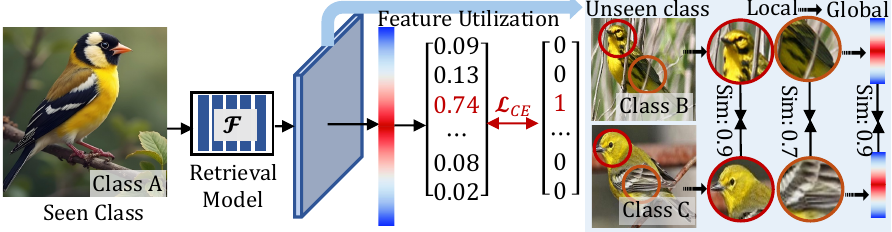}
         \caption{Category-guided Representation Learning}
         \label{1a}
     \end{subfigure}

     \begin{subfigure}[b]{1\linewidth}
         \centering
         \includegraphics[width=\linewidth]{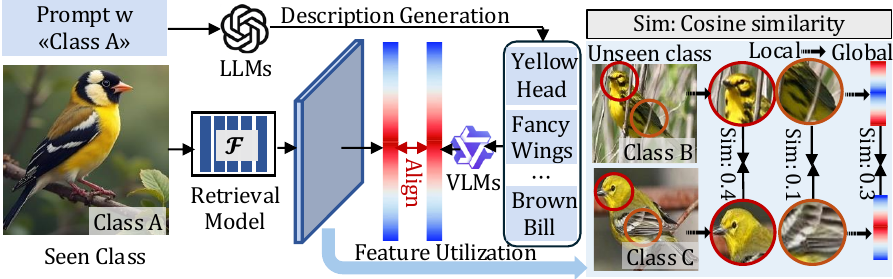}
         \caption{Language-driven Representation Learning}
         \label{1b}
         
     \end{subfigure}
     \setlength{\abovecaptionskip}{-0.15cm}
             \caption{
             Motivation of LaFG. (a) Learning with one-hot labels compresses class names into a single global identifiers and overlooks parts and attributes, making it hard to compare appearance details when facing unseen categories. Hence, similar local regions become indistinguishable, which degrades generalization to unseen categories. (b) Language-driven learning turns category names into linguistic supervision, thus establishing detail comparability. The model acquires transferable discriminative knowledge and improves retrieval on unseen categories.
        }
       
\end{figure}
Fine-grained image retrieval (FGIR) aims to retrieve visually similar images from the same subcategory, even for categories unseen during training. 
This capability is pivotal to real-world applications, such as fashion recommendation (e.g., fine-grained clothing retrieval~\cite{DBLP:conf/cvpr/LiuLQWT16,DBLP:conf/cvpr/AkKLT18}) and ecological monitoring (e.g., endangered-species recognition~\cite{DBLP:conf/cvpr/ElhoseinyZZE17, wei20212}). Motivated by its practical value, a large body of research has focused on learning representations that are both discriminative and generalizable to advance FGIR performance.

Recent studies on FGIR \cite{DBLP:conf/aaai/ParkPNK25,DBLP:conf/iclr/Ren0WH24,DBLP:conf/ijcai/LeW25} rely on supervisory signals derived from one-hot encodings of category names to retrieve visually similar subcategories. 
While effective on seen categories, such semantically sparse supervision fails to model comparability among cross-category details, hindering the acquisition of transferable discriminative details and ultimately limiting generalization to unseen categories. 
As illustrated in Fig.~\ref{1a}, visually similar parts (\textit{e.g.}, head or wing patterns) across unseen classes remain difficult to distinguish since one-hot supervision lacks part-aware guidance. Consequently, their local features collapse into similar representations, resulting in highly similar retrieval embeddings. This gap suggests a novel investigation: rather than compressing category information into one‑hot labels, can we unlock the semantics carried by category names and use them as supervisory signals that clearly reflects the comparability of object details?


This investigation aligns with vision–language models (VLMs) \cite{DBLP:conf/icml/RadfordKHRGASAM21,DBLP:conf/icml/JiaYXCPPLSLD21,DBLP:conf/cvpr/BendouOGB25} that unlock rich semantic priors from category names, naturally motivating language-driven supervision.
However, prevailing methods typically conceptualize category names as global identifiers, aligning entire images to them while consequently overlooking the comparability of object details.
Fortunately, recent advances in large language models (LLMs) \cite{DBLP:conf/acl/0009C23,DBLP:conf/nips/0001P00LSC24} make it feasible to generate detailed descriptions of object properties at scale, prompted by category names. 
Hence, we map LLM-generated descriptions into a vision-aligned embedding space via a frozen VLM and use these embeddings to supervise the retrieval model, thereby establishing detail comparability and facilitating generalization to unseen categories, as shown in Fig.~\ref{1b}.
Yet, raw LLM outputs are often incomplete, redundant, or noisy, making them inadequate for representing object appearances and thus providing unreliable supervision \cite{DBLP:journals/tois/HuangYMZFWCPFQL25, DBLP:conf/acl/LiCRCZNW24}. 
The key challenge, therefore, lies in designing a robust framework that automatically generates, refines, and aligns textual semantics with visual evidence, expanding category names into a more expressive and effective supervisory signal.

To this end, we introduce LaFG, a language-driven framework  for fine-grained retrieval that redefines a category name not as an index but as a semantic anchor, moving beyond the semantic narrowness of one-hot labels. 
We first prompt an LLM to unfold each name into diverse descriptions covering fine-grained attributes. 
These descriptions are mapped into a vision-aligned semantic space by a frozen VLM, and their embeddings are clustered across classes to induce a compact attribute vocabulary spanning the dataset.
The vocabulary serves two roles: denoising and completion—removing redundancy and noise in raw LLM outputs while borrowing complementary attributes from visually related categories. 
With a global prompt template, we query this vocabulary to retrieve a sparse attribute set and aggregate it into a class-specific linguistic prototype, yielding attribute-level targets that replace one-hot labels. 
Finally, these prototypes supervise the retrieval model to steer it toward pinpointing visual details consistent with linguistic descriptions within these prototypes, thus modeling comparability among object details. 

Our main contribution are summarized below:
\begin{itemize}
    \item 
    To the best of our knowledge, we are the first to expand category names beyond one-hot labels into semantically rich supervisory signals, thereby improving fine-grained generalization to unseen categories.
    
    \item
    We propose LaFG, a novel framework that implicitly model comparability between subtle object details via coupling LLMs and VLMs to automatically generate, refine, and align textual semantics with visual evidence.

    \item 
    Extensive experiments show that our LaFG achieves state-of-the-art performance on widely-used fine-grained and coarse-grained retrieval benchmarks.
\end{itemize}

\begin{figure*}[t]
\begin{center}
   \includegraphics[width=1\linewidth]{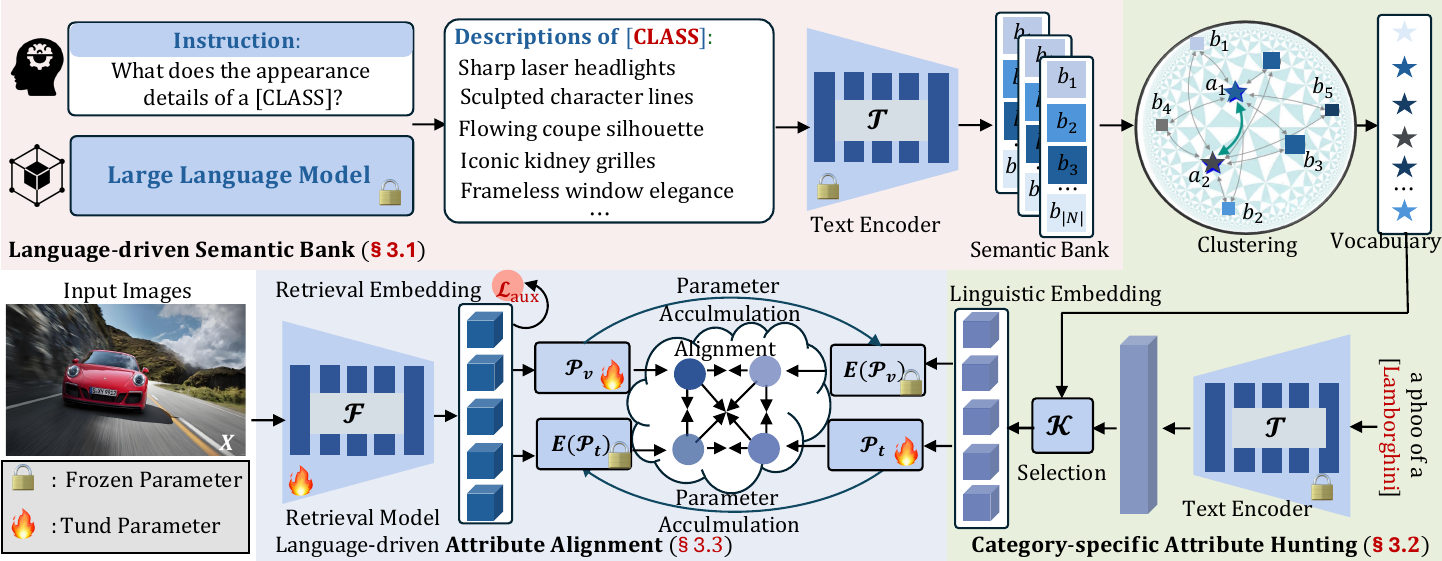}
\end{center}
\setlength{\abovecaptionskip}{-0.1cm} 
   \caption{Framework illustration of {
   \bf Language-driven Fine-grained Generalization}. See \S\ref{LaFG} for more details. }
\label{approach}
\end{figure*}

\section{Related Works}
\textbf{Fine-grained Image Retrieval.}
Recent advances in fine-grained image retrieval have primarily evolved along two distinct technical pathways \cite{DBLP:journals/tip/WeiLWZ17,DBLP:conf/ijcai/ZhengJSWHY18,DBLP:conf/wacv/MoskvyakMDB21, DBLP:conf/iccv/KoGK21,DBLP:conf/icml/RothMOCG21,DBLP:conf/iccv/ZhengZL021,DBLP:conf/cvpr/KimKCK21,DBLP:conf/cvpr/ZhengWL021}. Localization-based methods, exemplified by works like A$^2$-Net \cite{wei20212}, focus on precise object or part localization to enhance retrieval accuracy through reconstruction-based learning. Concurrently, metric-based approaches such as DDML~\cite{DBLP:conf/aaai/ParkPNK25} attempt to learn discriminative embedding spaces through sophisticated distance metrics, while NIA~\cite{DBLP:conf/cvpr/RothVA22} enforces unique translatability from class proxies, pulling same-subcategory samples closer in the embedding space.
However, such methods, which rely on supervisory signals derived from one-hot encodings of category names, struggles to capture comparability among cross-category details due to semantically sparse supervision, thereby limiting generalization to unseen categories.
To address this, our work, LaFG, extends category names beyond one-hot labels into semantically enriched supervisory signals, implicitly modeling comparability among object details and acquiring transferable discriminative details. 

\noindent\textbf{Vision-language Alignment.}
Vision–language alignment \cite{DBLP:conf/icml/RadfordKHRGASAM21,li2021supervision, jia2021scaling,tan2019lxmert} learns powerful joint representations across modalities by pre-training on large-scale image–text pairs. To mitigate modality discrepancy, existing approaches refine contrastive objectives to achieve more precise cross-modal alignment, either from the perspective of token-level correspondence \cite{DBLP:conf/iclr/YaoHHLNXLLJX22,wang2023tuning,bi2023vl} or multi-level semantic consistency \cite{DBLP:conf/eccv/Li0LZHZWH0WCG20, zhang2021vinvl,li2022mvptr}. Hence, these works provide more accurate and richer supervision signals of multiple granularity, thus semantically aligning vision and language. 
Unlike prior methods, we project visual embeddings into the linguistic prototype space induced by the VLM through distribution-wise maximum similarity learning.
In this way, LaFG steers the retrieval model toward pinpointing visual details consistent with linguistic descriptions within these prototypes, while simultaneously preserving instance-specific cues from the visual inputs. 

\noindent\textbf{Language-guided Learning.}
Language-guided learning has been widely applied across diverse domains, including global image editing \cite{DBLP:conf/cvpr/0005XXBDX21}, single-image reflection separation \cite{DBLP:conf/cvpr/ZhongHWLS24}, visual continual learning \cite{DBLP:conf/cvpr/Ni0ZHM0X24}, and semantic segmentation \cite{DBLP:conf/cvpr/HeJGR23}. They typically employ language modeling as a pretext task for visual learning \cite{DBLP:conf/cvpr/Desai021, DBLP:conf/eccv/SariyildizPL20}, using linguistic supervision from pre-trained language models to guide the learning of visual representations.
However, most existing methods adopt a unidirectional paradigm using static linguistic features as fixed supervision for visual extraction, but fail to account for scenarios where language descriptions are imprecise or incomplete.
In contrast, we construct an attribute vocabulary from LLM-generated descriptions and exploit the VLM’s shared vision–language space for noise-robust and complementary attribute selection.

\section{Methodology}
\label{LaFG}

The core of LaFG (Fig.~\ref{approach}) is to convert category names into semantically rich supervision. Given a category name, an LLM generates attribute-oriented descriptions, which a frozen VLM encodes to form a semantic bank. Cross-class clustering produces a compact, shared attribute vocabulary. A global prompt then selects a sparse, discriminative attribute set per category and aggregates it into a linguistic prototype that replaces one-hot labels. Finally, distribution alignment ties these prototypes to visual features, preserving instance-specific cues while enforcing language-consistent appearance modeling.


\subsection{Language-driven Semantic Bank}
One-hot labels provide limited semantic supervision, weakening cross-category fine-grained comparability. We address this by treating category names as semantic anchors, prompting an LLM to produce attribute-oriented descriptions, and encoding them with a VLM into a vision-aligned embedding space, thereby bridging the vision–language modality gap and forming a comprehensive semantic bank.

To align generic LLM knowledge with fine-grained visual semantics, we enrich subcategory names with fine-grained instructions. Specifically, we prompt the frozen LLM (\textit{e.g.}, GPT-4 \cite{DBLP:journals/corr/abs-2303-08774}) as follows:

\begin{quote}
    \makebox[\linewidth]{%
        \colorbox{gray}{%
            \hspace*{0mm} 
            \begin{minipage}{\dimexpr\linewidth+10\fboxsep\relax} 
                \fontsize{9pt}{10pt}\selectfont 
               ``Generate $\mathrm{n}$ distinct and descriptive statements that capture the key visual attributes of \texttt{[CLASS]}. Include holistic semantic characteristics and fine-grained textural details that would help distinguish \texttt{[CLASS]} from other visually similar subcategories. Texts should be of the form: an appearance description of \texttt{[CLASS]}. It + descriptive contexts".
            \end{minipage}%
        }%
    }
\end{quote}
Guided by this prompt, we obtain $\mathrm{n}$ descriptions $\mathrm{D_c}$ per class $\mathrm{c}$, each encoding rich visual semantics.

Next, we build a comprehensive semantic bank by leveraging a frozen pre-trained VLM text encoder $\Phi_\mathrm{t}(\cdot)$ (\textit{e.g.}, CLIP \cite{DBLP:conf/icml/RadfordKHRGASAM21}) to embed the generated descriptions into a unified semantic space.
This encoder possesses the unique ability to interpret language from a visual perspective, thereby mapping textual descriptions onto a vision-aligned feature manifold. 
Specifically, for each class $\mathrm{c}$, we encode its $\mathrm{n}$ descriptions $\mathrm{D_c}$ into a set of attribute embeddings:
\begin{equation}
\mathrm{B_c} = \left\{ \Phi_ \mathrm{t}(\mathrm{D_c^i}) \mid \mathrm{D_c^i} \in \mathrm{D_c} \right\},
\end{equation}
where $\mathrm{B_c} \in \mathbb{R}^{\mathrm{n\times d}}$ represents the set of $n$ attribute embeddings with the dimension $\mathrm{d}$ for class $\mathrm{c}$. 
This bank, denoted as $\mathcal{B} = [\mathrm{B_1}, \cdots, \mathrm{B_C}] \in \mathbb{R}^{\mathrm{C \times n \times d}}$, serves as a semantic repository that encapsulates visual semantics across all training categories, where $\mathrm{C}$ denotes the number of classes in the training set, and $\mathrm{d}$ is the dimension of embeddings. 

Unlike prior VLM-based approaches that directly rely on category names with limited fine-grained semantics, our method leverages the generative capability of LLMs to produce semantic-enriched descriptions, thereby enhancing generalization to unseen subcategories.


\subsection{Category-aware Attribute Hunting}
Leveraging these fine-grained semantics sourced from LLMs for representing fine-grained categories remains unreliable. Although LLMs can produce diverse descriptions under fine-grained prompts, they often miss portions of object appearances and introduce redundancy or noise with a high probability, primarily due to the lack of explicit fine-grained supervision during their pretraining.

Consequently, we refrain from directly fusing per-class texts from the semantic bank. Instead, we cluster all description embeddings across the training set to construct a global attribute vocabulary. The vocabulary serves two roles: denoising and completion—removing redundancy and noise in raw LLM outputs while borrowing complementary semantics from visually related categories. Given this vocabulary, we then use a simple hand-crafted retrieval template to hunt, for each category, the most semantically relevant attributes, yielding a compact yet expressive attribute set tailored to that category.

We consolidate the language-driven semantic bank $\mathcal{B}$ into a compact attribute vocabulary set $\mathcal{V}$. 
Concretely, we group embeddings into $|\mathrm{N}|$ universal attributes by applying K-means clustering $\mathcal{K}(\cdot)$ across all training categories:
\begin{equation}
\mathcal{V} = \mathcal{K}(\mathcal{B}, |\mathrm{N}|) = \{ \mathrm{a_i} \}_{\mathrm{i=1}}^{|\mathrm{N}|},
\end{equation}
where each cluster centroid $a_i \in \mathbb{R}^\mathrm{d}$ represents an attribute, a common semantic pattern that consistently emerge across multiple fine-grained descriptions. This clustering process automatically consolidates recurring semantic patterns and eliminates redundant descriptions, yielding a discriminative attribute vocabulary that effectively captures the essential visual traits shared across subcategories.

Building upon the universal attribute vocabulary $\mathcal{V}$, we perform category-aware attribute selection through a cross-text semantic alignment. For each class $\mathrm{c}$, we generate a hand-craft descriptor ``a photo of [\texttt{CLASS}]'' and encode it through the same CLIP's text encoder $\Phi_\mathrm{t}(\cdot)$ to obtain the category embedding $t_c \in \mathbb{R}^d$. 
Leveraging CLIP's shared vision-language embedding space, the category embedding encapsulates category-level semantic centroids \cite{DBLP:conf/icml/RadfordKHRGASAM21, DBLP:journals/ijcv/ZhouYLL22}. 
We thus employ it as a query to retrieve the most semantically relevant attributes from $\mathcal{V}$ based on their similarities. The selection process identifies the Top-$K$ attributes that maximize the similarity with the class prototype:
\begin{equation}
    \mathcal{V}_\mathrm{c} =\{ \mathrm{a_k:k}\in \argmax_{\mathrm{Top-K}} \{\mathrm{t_c^T \cdot a_k}\}_{\mathrm{k=1}}^{\mathrm{B}}\}.
    \label{eq3}
\end{equation}

We compute the final category prototype $\mathrm{T_c} \in \mathbb{R}^\mathrm{d}$ for class $\mathrm{c}$ by adaptively fusing its class embedding $t_c$ with the most relevant attributes $\mathcal{V}_c \in \mathbb{R}^{\mathrm{K\times d}}$. The representation is formulated as:
\begin{equation}
        \mathrm{T_c} = \mathrm{t_c} + \sum_{\mathrm{k=1}}^\mathrm{K} \sigma\!\big(\mathrm{t_c}^\top \mathrm{a_k}\big) \cdot \mathrm{a_k},
\label{eq4}
\end{equation}
where $\sigma(\cdot)$ applies softmax normalization to the similarity scores between $\mathrm{t_c}$ and each attribute $\mathrm{a_k} \in \mathcal{V}_\mathrm{c}$.

\subsection{Language-driven Attribute Alignment}

Aligning visual cues present in images with attribute-level representations derived from category names is a prerequisite for establishing comparability among fine-grained object details. 
To this end, we propose a language-driven attribute alignment module that supervise the retrieval model to steer it toward pinpointing visual details consistent with LLM-generated linguistic descriptions.


For an input image $\mathbf{X} \in \mathbb{R}^{3 \times H \times W}$, the retrieval model $\mathcal{F}$ extracts its representation as an embedding vector $\mathrm{V \in \mathbb{R}^{d}}$. Concurrently, we select the category prototype $\mathrm{T_c}$ corresponding to the image's category $\mathrm{c}$.
To supervise $\mathcal{F}$ towards learning visual features that align with linguistic descriptions, we first introduce two modality-specific linear projectors: $\mathcal{P}_\mathrm{v}$ for the visual embedding $\mathrm{V}$ and $\mathcal{P}_\mathrm{t}$ for the category-specific prototype $\mathrm{T_c}$, mapping their respective inputs into attribute-aligned embedding spaces.

Since each projector receives input from only one modality, it learns to model the attribute distribution specific to this modality. If both projectors produce the same distribution for a given embedding—regardless of its original modality—this indicates that the embedding has become modality-invariant, effectively supervise the retrieval model towards learning visual features that align with linguistic descriptions. 
This alignment is quantified by minimizing the symmetric Kullback-Leibler (KL) divergence
\begin{equation}
\begin{split}
   \hat{\mathcal{L}}_{\mathrm{ali}}  =  \mathrm{\mathcal{P}_\mathrm{v}(T_c|\theta_v)} & \log \mathrm{\frac{\mathcal{P}_\mathrm{v}(T_c|\theta_v)}{\mathcal{P}_\mathrm{t}(T_c|\theta_t)}}  \\ &+\mathrm{\mathcal{P}_\mathrm{t}(V|\theta_t) \log \frac{\mathcal{P}_\mathrm{t}(V|\theta_t)}{\mathcal{P}_\mathrm{v}(V|\theta_v)}},
        \label{eq5} 
\end{split}
\end{equation}
where $\mathrm{\theta_t}$ and $\mathrm{\theta_v}$ denotes the parameters of $\mathcal{P}_\mathrm{t}$ and $\mathcal{P}_\mathrm{v}$, respectively. 

However, we observe that direct optimization of this objective leads to premature convergence, where the projectors simply mimic each other's outputs without establishing meaningful attribute distribution alignment. To address this, we introduce mean projectors with exponentially moving average parameters that stabilize the training process:
\begin{equation}
\begin{split}
\mathrm{E^{(t)}[\theta_v]} &= \mathrm{(1-\alpha)\,E^{(t-1)}[\theta_v] + \alpha\,\theta_v,} \\
\mathrm{E^{(t)}[\theta_t]} &= \mathrm{(1-\alpha)\,E^{(t-1)}[\theta_t] + \alpha\,\theta_t,}
\end{split}
\label{eq6}
\end{equation}
where $\mathrm{ E^{(t)}[\theta]}$ and $\mathrm{E^{(t-1)}[\theta]}$ denote the parameters of mean projectors in the current iteration $\mathrm{t}$ and last iteration $\mathrm{t-1}$. The mean projectors are initialized as $\mathrm{ E^{(0)}[\theta_v]} = \mathrm{\theta_v}$ and $ \mathrm{E^{(0)}[\theta_t] = \theta_t}$. The parameter $\alpha$ is the updating ratio within the range of $(0,1]$.

In this way, Eq.~\eqref{eq5} can be rewritten as 
\begin{equation}
    \begin{split}
        \mathrm{\mathcal{L}_{ali}  = P_v(T_c|E[\theta_v]) } & \mathrm{\log \frac{P_v(T_c|E[\theta_v])}{P_t(T_c|\theta_v)} } \\ &\mathrm{+P_t(V|E[\theta_t]) \log \frac{P_t(V|[\theta_t])}{P_t(V|\theta_v)}}.
    \end{split}
\end{equation}
During training, since the two projectors are not updated via backpropagation, $\mathcal{L}_{\mathrm{ali}}$ exclusively optimizes the retrieval model by enforcing the distribution of visual embeddings to align with the corresponding category prototype. This alignment allows linguistic attributes to attend to diverse visual regions within images, effectively modeling comparability among fine-grained details and enhancing generalization to unseen categories. Crucially, LaFG aligns distributions instead of than individual embeddings, allowing retrieval embeddings to preserve instance-specific cues and stay consistent with linguistic descriptions.

\subsection{Overall Training Objective}
By transforming category names into semantically rich prototypes, the retrieval model is supervised to establish comparability among fine-grained visual details associated with each category.
To extend this comparability beyond individual instances, we introduce an auxiliary contrastive loss that encouraging the retrieval model to learn comparability among cross-category details.

During training, we sample $\mathrm{N}$ categories with two instances per class, resulting in a batch size of $\mathrm{K=2N}$.
For an anchor $\mathrm{z_i}$, the auxiliary contrastive loss is defined as:
\begin{equation}
\mathrm{\mathcal{L}_{aux}(z_i) = - \log \frac{\exp{(-D({z}_i, {z}_j) / \tau)}}{ \sum_{k=1, k \neq i}^K \exp{(-D({z}_i, {z}_k) / \tau) } }},
\end{equation}
where $\mathrm{(z_i, z_j)}$ denotes a positive pair from the same subcategory, and $\mathrm{(z_i, z_k)}$ represents all other sample pairs except the anchor itself.
Here, $\tau$ is a temperature hyperparameter, and $\mathrm{D(\cdot,\cdot)}$ represents the distance, implemented as the squared Euclidean distance between normalized vectors:
\begin{equation}
\mathrm{D({z}_i, {z}_j) = \norm{ \frac{{z}_i}{\norm{{z}_i}_2} - \frac{{z}_j}{\norm{{z}_j}_2} }^2_2 = 2 - 2 \frac{\langle{z}_i,{z}_j \rangle}{\norm{{z}_i}_2 \cdot \norm{{z}_j}_2}}.
\end{equation}

The total loss $\mathcal{L}$ of LaFG is formulated as
\begin{equation}
    \mathrm{\mathcal{L} = \mathcal{L}_{aux} + \beta \cdot \mathcal{L}_{ali}},
    \label{eq10}
\end{equation}
where $\beta$ is the hyper-parameter to balance the contributions of the individual loss item.

\section{Experiments}
\subsection{Experimental Setup}
\textbf{Datasets.} CUB-200-2011 \cite{Branson2014Bird} consists of 200 bird species. We use the first 100 subcategories (5,864 images) for training and the consists of (5,924 images) for testing. 
The Stanford Cars \cite{Krause20133D} includes 196 car models. Similarly, we use the first 98 classes, which contain 8,054 images, for training and the remaining classes, which contain 8,131 images, for testing. 
Stanford Online Products (SOP) \cite{DBLP:conf/cvpr/SchroffKP15} is divided into the 11, 318 subcategories (59, 551 images) in training, and the rest 11, 316 classes (60, 502 images) in testing. 
\textit{This split ensures \textbf{no category overlap} between training and testing sets, where all testing categories are strictly unseen during training to evaluate cross-category generalization.}

\noindent\textbf{Implementation Details.}
Our retrieval model is implemented on top of a Vision Transformer (ViT)~\cite{dosovitskiy2020image} initialized with ImageNet pre-trained weights. Input images are resized to $256 \times 256$ and randomly cropped to $224 \times 224$ during training. We adopt stochastic gradient descent with an initial learning rate of $1 \times 10^{-5}$, momentum of 0.9, and weight decay of $1 \times 10^{-4}$, using a batch size of 900 on NVIDIA A100 GPUs. To improve robustness, we apply standard data augmentations including random cropping, horizontal flipping, and color jittering. The learning rate follows an exponential decay schedule with a decay factor of 0.9 every five epochs over a total of 200 training epochs.

\noindent\textbf{Evaluation protocols.} We evaluate retrieval performance using Recall@K with cosine distance, following the standard protocol in prior work \cite{DBLP:conf/cvpr/SongXJS16}. Specifically, for each query image, the model retrieves the top-$M$ most similar images. A score of 1 is assigned if at least one positive image appears within the top-$M$ results, and 0 otherwise. The final Recall@K is obtained by averaging these scores across all queries in the test set.

\begin{table}\centering
	\caption{ Recall@1 performance comparison across constraint combinations on CUB-200-2011 benchmark.
	}
 \setlength{\tabcolsep}{12pt}
	\begin{tabular}{lll|l}
		\toprule[1pt]
		$\mathcal{L}_{aux}$ & $\hat{\mathcal{L}}_{ali}$ & $\mathcal{L}_{ali}$ & Recall@1 \\
		\toprule[0.7pt] 
	\checkmark&&&82.6\% \\
        \hline 
    \checkmark&\checkmark&&85.3\%$_{+2.7\%}$ \\
    &&\checkmark&86.5\%$_{+3.9\%}$ \\
    \rowcolor{gray}
    \checkmark&&\checkmark&\textbf{87.2\%}$_{+4.2\%}$\\
		\toprule[1pt]
		
	\end{tabular}\\
	\label{t1}
\end{table}

\begin{table}\centering
	\caption{ Evaluation results of retrieval performance on CUB-200-2011 with/without the synergy of LLMs and VLMs. ``Hand-crafted Language'' indicates the use of template-based textual descriptions (e.g., a photo of a [\(\cdot\)]) instead of LLM-generated descriptions.
	}
 \setlength{\tabcolsep}{12pt}
	\begin{tabular}{l|l}
		\toprule[1pt]
		Language & Recall@1\\
        \toprule[0.7pt]
        VLM + Hand-craft Language & 83.7\% \\
        VLM + LLM (w/o Vocabulary) & 85.3\%$_{+2.6\%}$ \\
        \rowcolor{gray}
        VLM + LLM & \textbf{87.2}\%$_{+3.5\%}$\\
		\toprule[1pt]
		
	\end{tabular}\\
	\label{t2}
\end{table}

\begin{table*}[!t]\centering
	\caption{ Compared with competitive methods on CUB-200-2011, Stanford Cars 196 and Stanford Online Products datasets. ``Arch'' denotes the backbone architecture. ``R50'' and ``ViT'' denote the ResNet-50~\cite{He2015Deep} and Vision Transformer~\cite{dosovitskiy2020image} backbones, respectively.
} 
	\begin{tabular}{l|l|cccc|cccc|cccc}
		\toprule[1pt]
       \multirow{2.5}{*}{Method} & \multirow{2.5}{*}{Arch}& \multicolumn{4}{c|}{CUB-200-2011} &\multicolumn{4}{c|}{Stanford Cars 196} & \multicolumn{4}{c}{Stanford Online Products}\\
		\cline{3-14}
		 &  & 1&2&4&8& 1&2&4&8& 1&10&100&1000 \\
         \toprule[0.7pt]
         CBML $_{\rm TPAMI23}$ \cite{DBLP:journals/pami/KanHCLMH23}&R50& 69.9 &80.4 &87.2 &92.5& 88.1 & 92.6 &95.4&97.4 &79.9 & 91.5&  96.5&98.9\\  
         NIR$_{\rm CVPR22}$ \cite{DBLP:conf/cvpr/RothVA22} &R50& 70.5 & 80.6 & -&-&89.1 & 93.4 &-&-&80.4 & 91.4& -& - \\
          HSE$_{\rm ICCV23}$ \cite{DBLP:conf/iccv/YangSLCCS23}& R50 & 70.6& 80.1& 87.1 & -&89.6& 93.8& 96.0& -&80.0 &91.4&96.3& - \\
          
IDML$_{\rm TPAMI24}$ \cite{DBLP:journals/pami/WangZZZL24} & R50 & 70.7& 80.2& - & -&90.6& 94.5& -& 81.5&-&-& -& - \\
HIST $_{\rm  CVPR{22}}$ \cite{DBLP:conf/cvpr/LimYP022} & R50& 71.4 &  81.1& 88.1&-&89.6& 93.9& 96.4&-&81.4 & 92.0& 96.7& - \\
PNCA++$_{\rm ECCV20}$ \cite{DBLP:conf/eccv/TehDT}& R50& 72.2 & 82.0 & 89.2 &93.5& 90.1& 94.5&97.0&98.4&81.4&  92.4& 96.9&99.0\\
			\toprule[0.7pt]
DIML$_{\rm TPAMI24}$ \cite{DBLP:journals/pami/ZhaoRZL24} & ViT & 76.7 & -&-&-& 80.7 &-&-&-& 79.5 &-&-&- \\
LM-Metric$_{\rm PR24}$ \cite{DBLP:journals/pr/YanXSLS24} & ViT & 77.1  & -&-&-& - &-&-&-& 83.1 &-&-&- \\
DFML-PA$_{\rm CVPR23}$ \cite{DBLP:conf/cvpr/WangZL0L23} & ViT & 79.1& 86.8& -& -& 89.5& 93.9& - &- &84.2& 93.8& -& -\\
DVA $_{\rm Arxiv25}$ & ViT & 84.9 &90.6 &94.5 &96.7& 90.7 &94.8& 97.1& 98.4& -& -& -& -\\
DPHM$_{\rm PR25}$ \cite{DBLP:journals/pr/XuCH25} & ViT& 85.5 & 91.3 & 94.6 & 96.6 & 84.1 & 90.5 & 94.5 & 97.1 & 84.8 & 94.5 & 98.1 & 99.4 \\
HypViT$_{\rm CVPR22}$ \cite{DBLP:conf/cvpr/ErmolovMKSO22} & ViT & 85.6 & 91.4 & 94.8& 96.7& 89.2 & 94.1 & 96.7 &98.1&85.9&94.9&98.1&99.5\\
HIER$_{\rm CVPR23}$ \cite{DBLP:conf/cvpr/KimJK23} & ViT & 85.7 & 91.3 & 94.4 & - & 88.3 & 93.2 & 96.1 & - & 86.1&95.0&98.0&-\\
SEE$_{\rm IJCAI25}$ \cite{DBLP:conf/ijcai/LeW25} &ViT &85.8 &91.4 &94.6& -& 88.8 &93.8 &96.4 &-&86.3 &95.0 &98.2& -\\
DDML$_{\rm AAAI25}$ \cite{DBLP:conf/aaai/ParkPNK25} & ViT & 86.0 & 91.7 & 95.2 & 96.8 & 89.5 & 94.2 & 96.8 & 98.2 & 86.1&95.1&98.2&99.5\\
VPTSP-G$_{\rm ICLR24}$ \cite{DBLP:conf/iclr/Ren0WH24} & ViT & 86.6 & 91.7 & 94.8 & - & 87.7 & 93.3 & 96.1 & - & 84.4&93.6&97.3&-\\
\toprule[0.7pt]
\rowcolor{gray}
Our LaFG & ViT & \textbf{87.2} & \textbf{92.4} & \textbf{95.2} & \textbf{97.0} & \textbf{91.5} & \textbf{94.6}& \textbf{96.6} & \textbf{98.5} & \textbf{87.1} &\textbf{ 95.8}& \textbf{98.5} & \textbf{99.5} \\

\toprule[1pt]
	\end{tabular}
    \label{t3}
\end{table*}

\subsection{Ablation Experiments}
\noindent\textbf{Efficacy of various constraints.} 
LaFG is optimized with a combination of two objectives: an auxiliary loss $\mathrm{\mathcal{L}_{aux}}$ and an alignment loss $\mathrm{\mathcal{L}_{ali}}$ (or its variant $\mathrm{\hat{\mathcal{L}}_{ali}}$). These losses serve complementary roles to effectively convert category names into semantically rich supervision, thus establishing the comparability of cross-category details.
Ablation studies on CUB-200-2011 (Tab.~\ref{t1}) reveal critical insights: Training solely with $\mathcal{L}_{\text{aux}}$ achieves 82.6\% Recall@1 but fails to model detail comparability, limiting generalization to unseen categories. 
Incorporating $\hat{\mathcal{L}}_{\text{ali}}$ boosts Recall@1 to 85.3\%, demonstrating that LLM-generated descriptions provide valuable signals. 
However, by optimizing projectors to mimic the distributions of category-specific prototypes rather than directly refining visual representations, $\hat{\mathcal{L}}_{\text{ali}}$ remains insufficient. To overcome this, we introduce exponential moving average (EMA) updates replacing backpropagation. This innovation enables $\mathcal{L}_{\text{ali}}$ to directly optimize the retrieval model sensitive to semantic comparison. Upon this, the group of $\mathcal{L}_{\text{ali}}$ and $\mathcal{L}_{\text{aux}}$ successfully establishes comparability among cross-category details, thus acquiring state-of-the-art 87.2\% Recall@1 on the CUB-200-2011 benchmark.

\noindent\textbf{Importance of the synergy of LLMs and VLMs.}
We assess the synergy between LLMs and VLMs by varying both the source and the processing of textual descriptions (Tab.~\ref{t2}). Using handcrafted CLIP prompts (“a photo of a [$\cdot$]”) as a baseline yields only marginal gains, since such templates lack the fine-grained cues contributing to decision boundary. Replacing them with LLM-generated, attribute-centric descriptions injects complementary semantics and raises Recall@1 to 85.3\%. However, raw LLM texts remain incomplete and noisy. To consolidate and share attributes across related subcategories, we cluster VLM text embeddings of all descriptions to induce a compact, dataset-wide attribute vocabulary, then retrieve a relevant subset per category to form a category-specific prototype. These prototypes supervise the retrieval model to attend to attribute-consistent visual details, further boosting accuracy to 87.2\% and underscoring the effectiveness of coupling LLMs with VLMs.

\subsection{Comparison with State-of-the-art Methods}
\noindent\textbf{Fine-grained Image Retrieval.}
Our LaFG establishes new state-of-the-art performance on both zero-shot fine-grained image retrieval benchmarks (CUB-200-2011 and Stanford Cars-196), achieving Recall@1 accuracies of 87.2\% and 91.5\% respectively, as detailed in Tab.~\ref{t3}. We achieve absolute Recall@1 gains over SEE  \cite{DBLP:conf/ijcai/LeW25} (+1.4\% on CUB, +2.7\% on Cars) and DDML  \cite{DBLP:conf/aaai/ParkPNK25} (+1.2\% on CUB, +2.0\% on Cars), highlighting the advantage of converting category names from one-hot labels to seamntically rich supervisory signals.
Furthermore, prior work such as VPTSP-G~\cite{DBLP:conf/iclr/Ren0WH24} typically exploits the zero-shot capability of foundation models to learn discriminative and generalizable embeddings. In contrast, LaFG employs a frozen VLM to project LLM-generated descriptions into a vision-aligned semantic space, induces category-specific attribute prototypes, and uses these prototypes to supervise the retrieval model, guiding it to localize visual details consistent with language descriptions and yielding superior performance.
The consistent improvements substantiate that converting class names into attribute-level supervision explicitly models cross-category comparability, thereby enhancing generalization to unseen categories.

\noindent\textbf{Coarse-grained Image Retrieval.}
To assess LaFG's generalization capacity beyond fine-grained domains, we conduct a large-scale evaluation on the coarse-grained benchmark, \textit{i.e.}, Stanford Online Products, as reported in Tab.~\ref{t3}. The framework synergistically leverages LLMs and VLMs to project LLM-generated descriptions into a vision-aligned space via VLMs, enabling accurate representation of both seen and unseen categories.
This synergy enables LaFG to resolve subtle fine‑grained distinctions and to bridge coarse‑grained semantic gaps, yielding superior performance across levels of granularity. Consistent improvements on SOP further confirm that converting class names into attribute‑level supervision scales effectively from fine‑grained to coarse‑grained retrieval.

\begin{table}\centering
	\caption{ Retrieval performance on CUB-200-2011 for models trained with different numbers of language descriptions $n$ generated by LLMs.
	}
 \setlength{\tabcolsep}{5pt}
	\begin{tabular}{l|ccccc}
		\toprule[1pt]
		Number $n$ & 5 &10 &15 &20&25\\
        \toprule[0.7pt]
        Recall@1 &81.4\% &84.3\% &85.7\% &87.2\% &86.3\%\\
		\toprule[1pt]
	\end{tabular}\\
	\label{t4}
\end{table}

\begin{table}\centering
	\caption{ Retrieval performance on CUB-200-2011 for models trained with different attribute vocabulary sizes $\mathrm{|N|}$.
	}
 \setlength{\tabcolsep}{5pt}
	\begin{tabular}{l|ccccc}
		\toprule[1pt]
		Size $\mathrm{|N|}$ & 32 &64 &128 &256&512\\
        \toprule[0.7pt]
        Recall@1 &84.3\% &85.7\% &87.2\% &86.3\% &85.9\%\\
		\toprule[1pt]
	\end{tabular}\\
	\label{t5}
\end{table}

\begin{table}\centering
	\caption{ Evaluation on CUB-200-2011 for models trained with different Top-$K$ category-specific attributes used to represent each training category in Eq.~\eqref{eq3}.
	}
 \setlength{\tabcolsep}{5pt}
	\begin{tabular}{l|ccccc}
		\toprule[1pt]
		Top-K & 16 & 24 &32 &40&48\\
        \toprule[0.7pt]
        Recall@1 &84.1\% &85.2\% &86.3\% &87.2\% &86.9\%\\
		\toprule[1pt]
		
	\end{tabular}\\
	\label{t6}
\end{table}

\subsection{Discussion}
\noindent\textbf{Effect on different numbers of language descriptions.} 
As shown in Tab.~\ref{t4}, retrieval performance on CUB-200-2011 exhibits a clear dependence on the number of language descriptions ($n$) generated by LLMs. Recall@1 steadily increases with larger $n$, peaking at 87.2\% when $n=20$, before slightly decreasing to 86.3\% at $n=25$. This non-monotonic behavior reveals two competing effects: when $n \leq 10$, insufficient linguistic cues fail to capture fine-grained visual distinctions, resulting in suboptimal performance (81.4\%–84.3\%); when $n > 20$, redundant or noisy descriptions begin to obscure discriminative attributes, causing a 1.1\% drop from $n=20$ to $n=25$. Overall, using around 20 high-quality descriptions achieves the best trade-off between semantic richness and precision for fine-grained representation learning.

\noindent\textbf{Investigation on the size of attribute vocabulary.}
The results in Tab.~\ref{t5} demonstrate a non-linear relationship between attribute vocabulary size ($\mathrm{|N|}$) and retrieval performance. Recall@1 peaks at 87.2\% when $\mathrm{|N|}=128$, while both smaller and larger vocabularies yield reduced effectiveness. With a small vocabulary size (e.g., $\mathrm{|N|}=32$), multiple distinctive attributes collapse into a single cluster, leading to the loss of discriminative details. In contrast, an excessively large vocabulary  (e.g., $\mathrm{|N|}=512$) causes over-segmentation, where semantically similar descriptions are divided into separate attributes, introducing redundancy and noise that degrade performance. The optimal $\mathrm{|N|}=128$ strikes an effective balance between preserving fine-grained distinctiveness and maintaining attribute robustness.
\begin{table}\centering
	\caption{ Quantitative performance of the model on CUB-200-2011 when trained with different dimension of  distribution in the two projectors.
	}
 \setlength{\tabcolsep}{5pt}
	\begin{tabular}{l|ccccc}
		\toprule[1pt]
		Dimension & 32 &64 &128 &256 &512\\
        \toprule[0.7pt]
        Recall@1 &80.1\% &84.5\% &86.0\% &87.2\% &86.9\%\\
		\toprule[1pt]
		
	\end{tabular}\\
	\label{t8}
\end{table}
\begin{table}\centering
	\caption{ Evaluation on CUB-200-2011 for models trained with different update ratios $\alpha$ in the parameter updating defined in Eq.~\eqref{eq6}.
	}
 \setlength{\tabcolsep}{5pt}
	\begin{tabular}{l|ccccc}
		\toprule[1pt]
		Ratio $\alpha$ & 0.1 &0.2 &0.4 &0.6&0.8\\
        \toprule[0.7pt]
        Recall@1 &87.1\% &87.2\% &86.9\% &86.6\% &86.6\%\\
		\toprule[1pt]
		
	\end{tabular}\\
	\label{t7}
\end{table}

\begin{table}\centering
	\caption{ Quantitative performance of the model on CUB-200-2011 when trained with different weights $\beta$ in the loss function defined in Eq.~\eqref{eq10}.
	}
 \setlength{\tabcolsep}{5pt}
	\begin{tabular}{l|ccccc}
		\toprule[1pt]
		Weight $\beta$ & 1 &5 &10 &15 &20\\
        \toprule[0.7pt]
        Recall@1 &83.5\% &86.3\% &87.2\% &87.1\% &86.5\%\\
		\toprule[1pt]
		
	\end{tabular}\\
	\label{t9}
\end{table}

\noindent\textbf{How many attributes effectively represent objects?}
Tab.~\ref{t6} presents the impact of Top-$\mathrm{K}$ attribute selection on retrieval performance. Recall@1 increases with larger $\mathrm{K}$, peaking at 87.2\% when $\mathrm{K}=40$. Specifically, when using a smaller $\mathrm{K}$, the limited number of attributes fails to capture the full range of visual cues within each category, resulting in incomplete semantic representation. In contrast, larger $\mathrm{K}$ values introduce irrelevant attributes from other categories, reducing discriminative power. The optimal setting ($\mathrm{K}=40$) strikes the best trade-off, capturing meaningful attributes while remaining selective enough to preserve inter-class distinctiveness. The minor 0.3\% drop at $\mathrm{K}=48$ indicates slight attribute confusion, emphasizing the need for careful attribute selection in fine-grained retrieval.

\noindent\textbf{Effect on dimension of attribution distribution.}
In Tab.~\ref{t8}, retrieval accuracy improves as the prototype distribution dimension increases from 32 to 256, indicating that higher capacity better captures the rich, fine-grained attributes needed to highlight subtle inter-class differences. However, pushing the dimension further to 512 slightly degrades performance, likely because an over-parameterized prototype space encourages visual features to overfit class-specific prototypes and overlook instance-specific cues. In contrast, a small dimension under-represents attribute diversity and limits the model’s ability to obtain important distinctions.

\noindent\textbf{Effect on parameter accumulation.}
Tab.~\ref{t7} illustrates the effect of the exponentially moving average (EMA) update ratio ($\alpha$) on model performance, highlighting key insights into parameter optimization dynamics. 
The best performance (87.2\% Recall@1) at $\alpha=0.2$ indicates an optimal balance between parameter stability, which preserves sufficient historical information, and model adaptability, which incorporates new discriminative cues. 
A smaller ratio ($\alpha=0.1$) yields slightly lower performance due to over-reliance on historical parameters, which slows the integration of new attribute representations. 
Conversely, higher ratios ($\alpha \geq 0.4$) bias updates toward recent parameters, causing unstable training and noisy attribute projections. 
Overall, moderate EMA updates strike a balance between stability and adaptability.

\noindent\textbf{Hyperparameter analysis.}
In Tab.~\ref{t9}, the loss weight $\beta$ critically governs how effectively the model exploits implicit attribute supervision and enforces cross‑category attribute comparability. Performance peaks at $\beta$=10, indicating a balanced interplay between the language‑driven alignment loss and the auxiliary contrastive loss. Underweighting  weakens attention to language‑guided visual cues, whereas overweighting over-regularizes toward linguistic descriptions, thus degrading FGIR performance.

\begin{figure}[t]
\begin{center}
   \includegraphics[width=1\linewidth]{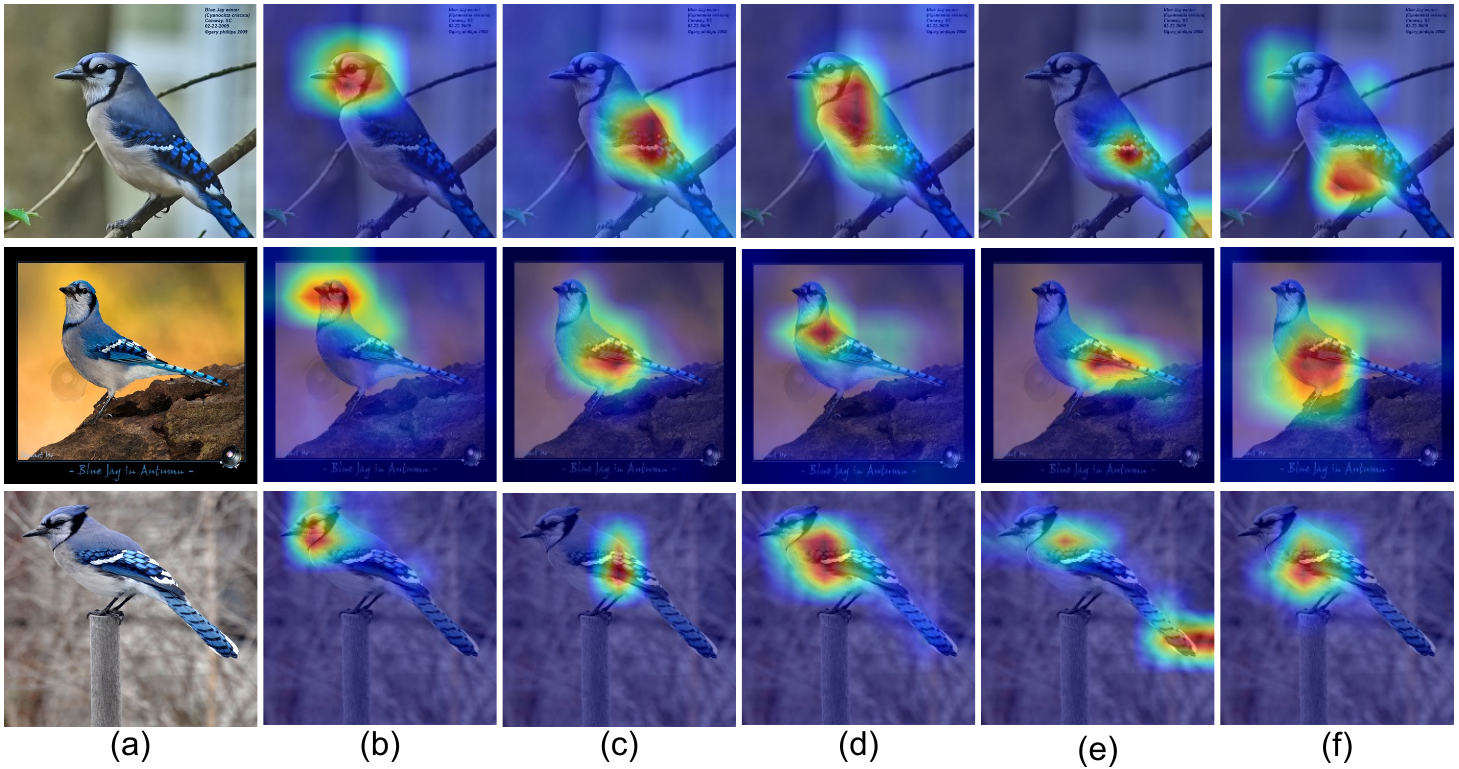}
\end{center}
\setlength{\abovecaptionskip}{-0.2cm} 
   \caption{
   Visualization of clustered attribute responses for the same subcategory (e.g., Blue Jay). The Top-5 attributes are selected from the vocabulary based on their similarity scores. (a) Input images; (b)–(f) Attribute response regions for comparison.
   }
   \label{attribute}
\end{figure}

\noindent\textbf{Visual attribute analysis.}
To evaluate whether the constructed attribute vocabulary effectively captures discriminative object characteristics, we visualize the top-5 clustered attributes ranked by similarity scores (Fig.~\ref{attribute}), using the Blue Jay category as an example. Two key observations emerge: each attribute consistently highlights salient regions (e.g., head, neck, wings), indicating that the learned vocabulary aligns with coherent visual semantics. Moreover, different attributes focus on diverse visual cues (e.g., beak shape vs. wing pattern), demonstrating their complementary roles in fine-grained characterization.
These findings confirm that LaFG encodes semantically meaningful attributes that correspond to localized visual features, provide diverse yet coherent appearance descriptors.

\begin{figure}[t]
\begin{center}
   \includegraphics[width=1\linewidth]{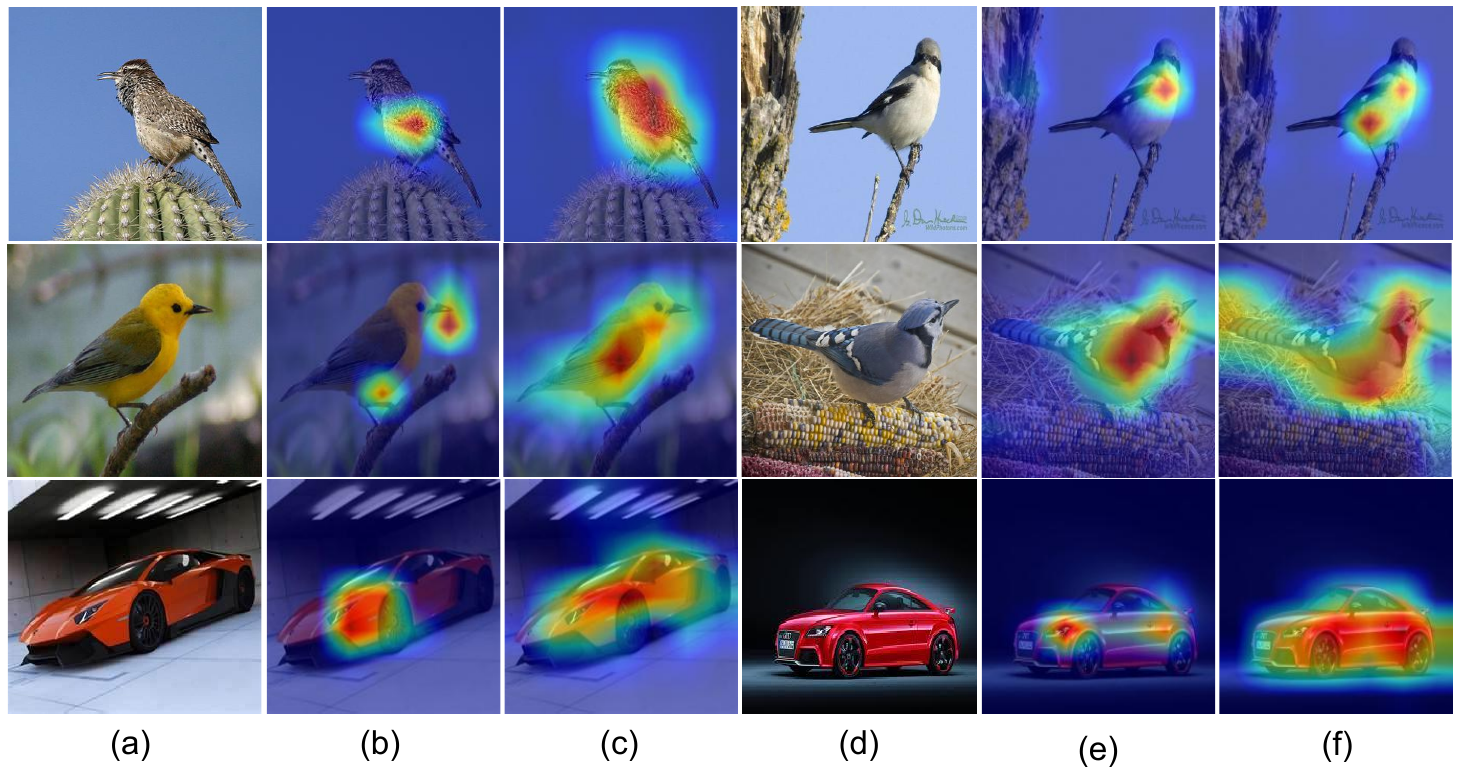}
\end{center}
\setlength{\abovecaptionskip}{-0.2cm} 
   \caption{Illustration of class activation maps generated by the baseline and our LaFG. (a) and (d) show the input images; (b) and (e) present the corresponding class activation maps produced by the baseline; (c) and (f) display the maps generated by our LaFG.
   }
   \label{activation}
\end{figure}

\noindent\textbf{Feature representation analysis.}
As shown in Fig.~\ref{activation}, the class activation maps demonstrate how language descriptions enhance visual representation learning. Compared with the baseline, which narrowly focuses on the most discriminative regions, LaFG exhibits two notable advantages. First, it activates additional informative regions (e.g., subtle texture patterns) that the baseline overlooks, preserving features essential for recognizing unseen categories. Second, language guidance directs the model to attend to fine-grained appearance details (e.g., feather patterns) rather than coarse category-level semantics, thereby improving generalization. These visualizations demonstrate that language-guided training supplies complementary visual cues absent in conventional works, thereby expanding and refining the model’s visual understanding and explaining LaFG’s superior generalization performance.

\section{Conclusion}
This paper introduces LaFG, a language-driven framework for FGIR that redefines a category name not as an index but as a semantic anchor, moving beyond the semantic narrowness of one-hot labels. 
LaFG implicitly models comparability among object details and acquires transferable discriminative details, thus improving generalization to unseen categories.
Comprehensive evaluations on fine‑grained and coarse‑grained benchmarks confirm that turning class names beyond one‑hot labels into attribute‑level supervisory signals provides large gains and strong generalization across seen and unseen categories.

{
    \small
    \bibliographystyle{ieeenat_fullname}
    \bibliography{main}
}

\end{document}